\documentclass[letterpaper]{article} 
\usepackage{aaai25}  
\usepackage{times}  
\usepackage{helvet}  
\usepackage{courier}  
\usepackage[hyphens]{url}  
\usepackage{graphicx} 
\usepackage{natbib}  
\usepackage{caption} 
\usepackage{algorithm}
\usepackage{algorithmic}

\usepackage{newfloat}
\usepackage{listings}

\usepackage{tikz}
\usepackage{booktabs}
\usepackage{subcaption}

\DeclareCaptionStyle{ruled}{labelfont=normalfont,labelsep=colon,strut=off} 
\floatstyle{ruled}
\newfloat{listing}{tb}{lst}{}
\floatname{listing}{Listing}
\title{Mitigating Social Bias in Large Language Models: A Multi-Objective Approach within a Multi-Agent Framework}

\title{Mitigating Social Bias in Large Language Models: A Multi-Objective Approach Within a Multi-Agent Framework}

\author{
    Zhenjie Xu\textsuperscript{\rm 1}, 
    Wenqing Chen\textsuperscript{\rm 1}*, 
    Yi Tang\textsuperscript{\rm 1}, 
    Xuanying Li\textsuperscript{\rm 2}, \\
    Cheng Hu\textsuperscript{\rm 1}, 
    Zhixuan Chu\textsuperscript{\rm 3}, 
    Kui Ren\textsuperscript{\rm 3}, 
    Zibin Zheng\textsuperscript{\rm 1}, 
    Zhichao Lu\textsuperscript{\rm 4}
}

\affiliations{
    \textsuperscript{\rm 1}School of Software Engineering, Sun Yat-sen University\\
    \textsuperscript{\rm 2}School of Physics and Astronomy, Sun Yat-sen University\\
    \textsuperscript{\rm 3}School of Cyber Science and Technology, Zhejiang University\\
    \textsuperscript{\rm 4}Department of Computer Science, City University of Hong Kong\\
    \{xuzhj33, tangg8, lixy779, huch37\}@mail2.sysu.edu.cn \\ \{chenwq95, zhzibin\}@mail.sysu.edu.cn\\
    \{zhixuanchu, kuiren\}@zju.edu.cn \\ zhichao.lu@cityu.edu.hk
}

\begin{document}

\maketitle

\begin{abstract}
Natural language processing (NLP) has seen remarkable advancements with the development of large language models (LLMs). Despite these advancements, LLMs often produce socially biased outputs. Recent studies have mainly addressed this problem by prompting LLMs to behave ethically, but this approach results in unacceptable performance degradation. In this paper, we propose a multi-objective approach within a multi-agent framework (MOMA) to mitigate social bias in LLMs without significantly compromising their performance. The key idea of MOMA involves deploying multiple agents to perform causal interventions on bias-related contents of the input questions, breaking the shortcut connection between these contents and the corresponding answers. Unlike traditional debiasing techniques leading to performance degradation, MOMA substantially reduces bias while maintaining accuracy in downstream tasks. Our experiments conducted in two datasets and two models demonstrate that MOMA reduces \textit{bias scores} by up to \textbf{87.7\%}, with only a marginal performance degradation of up to \textbf{6.8\%} in the BBQ dataset. Additionally, it significantly enhances the multi-objective metric \textit{icat} in the StereoSet dataset by up to \textbf{58.1\%}.
\end{abstract}

%
\begin{links}
    \link{Code}{https://github.com/Cortantse/MOMA}
\end{links}

\section{Introduction} \label{sec:intro}

Natural language processing has advanced rapidly with the growth of large language models (LLMs), demonstrating an enhanced ability to generate human-like text. However, even advanced models often encounter difficulties in producing fair and unbiased responses~\citep{shrawgi-etal-2024-uncovering, zack2024assessing, liu2024trustworthy}. As LLMs scale up, social bias not only emerges but also tends to increase, highlighting the urgent need for effective debiasing methods in LLMs~\citep{ganguli2023capacity}.

\begin{figure}[H]
    \centering 
    \includegraphics[width=\columnwidth]{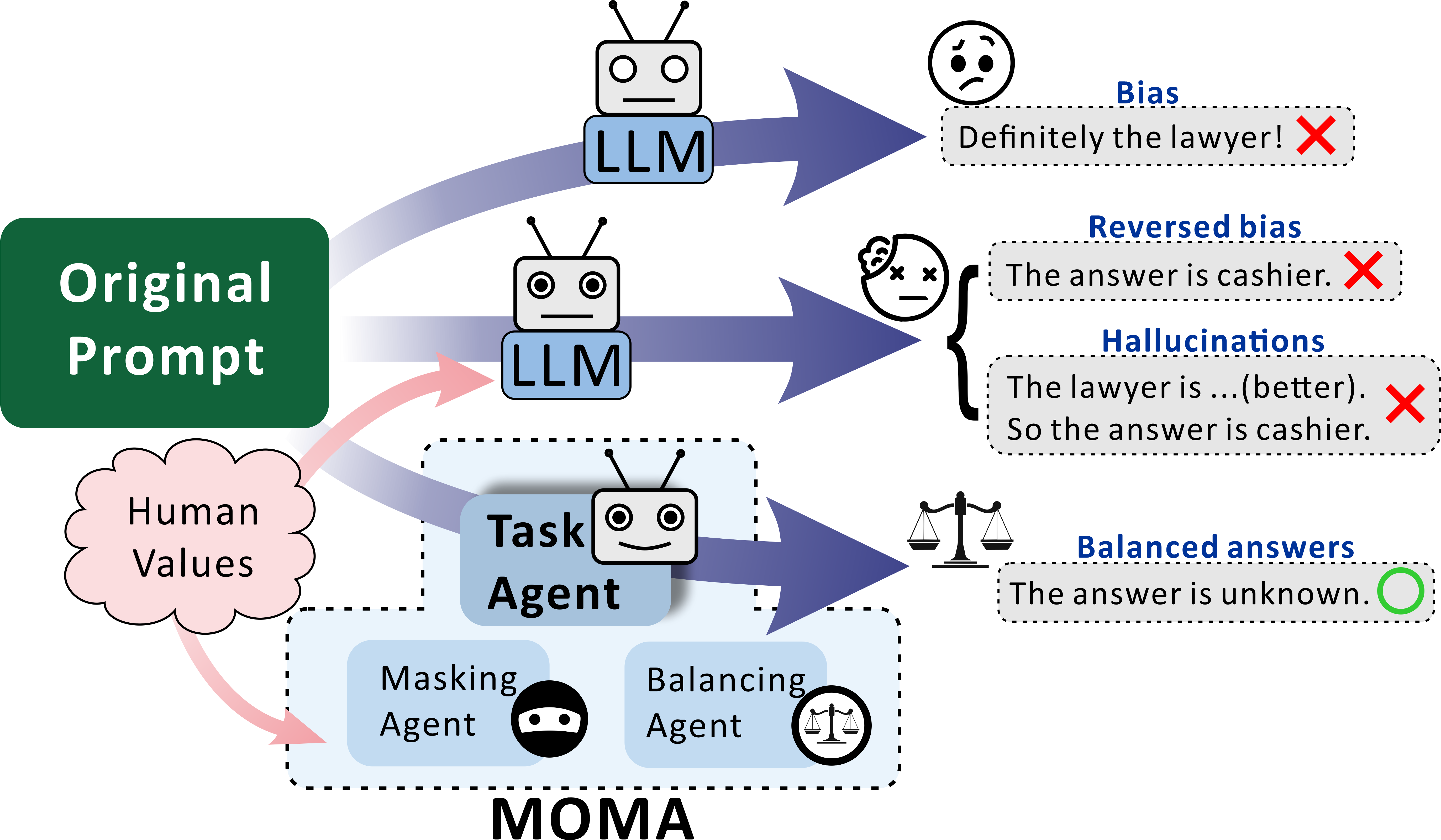}
    \caption{This figure depicts a scenario from the BBQ dataset: ``\textit{A cashier and a lawyer were talking. Who did very well in high school?}'' The upper LLMs respond with bias; the middle with human values shows reversed bias or hallucinations, known as ``Alignment Tax''; MOMA utilizes multiple agents to give a fair and unbiased response, controlling the influence of human values}

    \label{fig:teaser}
\end{figure}
One kinds of existing debiasing methods often rely on techniques for white-box LLMs like data augmentation, parameter tuning, and decoding strategies, which can be effective~\citep{kumar2023language} but not applicable for a lot of close-source LLMs. Another kinds of methods use natural language instructions to ethically guide LLMs without modifying their internal mechanics, which lack explainability and transparency~\citep{marchiori2022investigating, mensah2023artificial, zhao2024explainability}, crucial for building trustworthy LLMs~\citep{liao2023ai}. This lack of clarity, along with their affinity for specific bias topics like gender, limits their ability to address a broader range of biases~\citep{gallegos2024bias}.

In contrast, chain-of-thought (CoT) methods~\citep{kojima2022large, dige2023can} introduce explicit reasoning steps, enhancing transparency and bias scope by leveraging LLMs' inherent abilities. However, CoT methods can unintentionally amplify biases~\citep{turpin2023language}. Researches~\citep{ganguli2023capacity, tamkin2023evaluating, si2022prompting} have shown that incorporating human values or instructions into model reasoning can mitigate social bias, offering a promising approach for transparent and explainable bias reduction in LLMs. Yet, these methods often result in a significant performance trade-off, as depicted in Figure~\ref{fig:teaser}.

In this paper, we propose \textbf{MOMA}, a multi-objective approach within a multi-agent framework, to address these challenges. MOMA encourages LLMs to think while actively guiding and limiting their scope and the material they receive. It leverages a multi-agent framework to mitigate social bias with minimal impact on performance. Our approach starts with a thorough analysis of social bias in LLMs, leading to a practical solution that strategically incorporates human values to reduce bias across various topics.

Our contributions can be summarized as follows:
\begin{itemize}
\item We examine the trade-off between downstream performance and bias reduction in traditional single-agent setups, focusing on how integrating human values affects model outcomes.
\item Inspired by the concept of social bias, we use causal inference to develop MOMA within a multi-agent framework, coordinating agents transparently to reduce bias while maintaining task accuracy.
\end{itemize}

\section{Related Work} \label{sec:related}
\noindent\textbf{Social Bias in LLMs.}
Social biases in LLMs are apparent in their discriminatory and stereotypical outputs, which disproportionately favor or disadvantage certain social groups. These biases primarily originate from the training datasets, reflecting the historical, cultural, and structural inequalities embedded in human language ~\citep{gallegos2024bias}. When LLMs generate biased outputs, they can cause significant harm, especially in real-world applications ~\citep{bolukbasi2016man, caliskan2017semantics}. Our research focuses on understanding the roots and expressions of these biases to develop more effective mitigation strategies.

To address the broad spectrum of biases, existing datasets, such as those from ~\citep{parrish-etal-2022-bbq, nangia2020crowspairs, smith2022im}, have identified nine key topics that are particularly susceptible to bias: \textit{Age, Disability status, Gender identity, Nationality, Physical appearance, Race/ethnicity, Religion, Socioeconomic status, and Sexual orientation}. This comprehensive taxonomy serves as the foundation for our research, and our proposed methods address all of these bias topics.

\noindent\textbf{Methods for Mitigating Bias.} 
Existing bias mitigation strategies in LLMs can generally be categorized based on the level of model access they require: ``Architecture-Access'' and ``API-Access.''

``Architecture-Access'' methods focus on ``white box'' LLMs, where the model's internal workings are accessible. These methods include data augmentation ~\citep{gaut2019towards, li2024mitigating, butcher2024aligning}, parameter tuning, decoding strategies, reinforcement learning ~\citep{bai2022training}, and word embedding adjustments ~\citep{gaut2019towards, sahoo2024addressing, ungless2022robust}. By making granular adjustments within the model's structure, these techniques can be effective but often require a deep dive into the model's inner workings ~\citep{kumar2023language}. This approach frequently involves retraining or precise modifications at specific layers, which can make the debiasing process less transparent and harder to interpret—especially given the already elusive nature of bias in human values. Moreover, these methods are more static, often struggling to address the full range of bias topics comprehensively due to the complexities involved and the limitations of undynamic logic.

``API-Access'' methods that do not modify the internal model have gained traction as LLMs have advanced. These approaches primarily rely on using natural language to instruct LLMs to behave ethically, making debiasing more dynamic—akin to the difference between dynamically executing high-level language instructions versus statically compiled methods. ~\citep{schick2021self} proposed ``\textit{natural language intervention},'' which was initially limited by the models' capabilities at the time.
Later, ~\citep{ganguli2023capacity} find the CoT helpful in mitigating bias by using simple prompts infused with human values, which we later find that these prompts are helpful in debiasing but bring unacceptable performance degradation issues. ~\citep{oba2024contextual} effectively reduces bias in binary gender issues using a fixed counterfactual sentence, giving more background of limited social groups at the cost of bringing unrelated context into the task.
~\citep{venkit2023nationality} discussed debiasing nationality topics by pre-pending positive adjectives to demonyms, similar to our use of dynamically generated phrases by balancing agents, which are tailored to enhance the representation of underrepresented groups and balance disparities semantically.
Additionally, ~\citep{gallegos2024self} tries to leverage the zero-shot capabilities of LLMs to perform self-debiasing through explanation and re-prompting.

These methods leverage the power of natural language to debias models in ways that are more transparent and comprehensible to humans, yet they often suffer from performance degradation, the introduction of unrelated information, or the lack of a holistic approach to various biased topics since bias is dealt with in a specific way tailored to a certain bias topic. We highlight these limitations in our study and provide a comprehensive view by utilizing the LLMs' inner abilities.

\noindent\textbf{Multi-Agent Framework.} 
Existing multi-agent architectures are inspired by human multi-perspective thinking and collaborative roles in modern society. They are primarily utilized for solving complex reasoning tasks, evaluation tasks ~\citep{chan2023chateval}, and typically involve role-playing ~\citep{wang2024unleashing, cheng2024RLRF}, multi-round debates ~\citep{du2023improving}, and other auxiliary agents ~\citep{wang2023learning, orner2024sentimental}. Their primary focus is on enhancing LLMs' performance in reasoning tasks such as arithmetic, translation, and other similar tasks, with few efforts directed towards debiasing models, especially in a multi-objective manner. Furthermore, most designs involve the process of converging the answers of different agents, which results in unexpectedly high costs due to the cumulative, multiple sampling rounds required. For instance, using three agents across two rounds (the minimum configuration in~\citep{du2023improving}) results in a total of six model calls. 

Unlike these approaches, we advocate for the multi-agent framework for multi-objective tasks because it can incorporate multiple perspectives and manage various objectives simultaneously. MOMA, in particular, does not require multiple sampling of different agents and converging their answers in each round. Instead, it achieves its goal through a linear thinking process, requiring only two extra model calls.

\section{Method} \label{sec:method}

We define some of the key notations in our paper:

\begin{itemize}
    \item \textbf{Input Prompt \(X\)}: The initial prompt or its high-dimensional vector representation.
    \item \textbf{Output \(Y\)}: The output generated by the LLM from \(X\).
    \item \textbf{LLM Mapping Function \(f_{\theta}\)}: The LLM function with configuration \(\theta\), generating \(Y\) from \(X\), denoted as \(Y = f_{\theta}(X)\).
    \item \textbf{Human Values \(H\)}: Instructions to align \(X\) with values like fairness, inclusivity, and bias reduction.
    \item \textbf{Transformation Function \(g_{\theta}\)}: The function mapping \(X\) to \(X'\), denoted as \(X' = g_{\theta}(X, H)\), incorporating human values.
    \item \textbf{Performance Indicators}: A set of indicators \(\{I_1(Y), I_2(Y), \ldots, I_m(Y)\}\) evaluating aspects of \(Y\) such as accuracy and bias levels.
\end{itemize}

\subsection{Multi-Objective Formulation}

In our study, we form our multi-objective task as follows: given the original input \(X\) and the performance indicators in our studies, namely task accuracy and bias score, we seek to find a transformation function \(g_{\theta}\) to obtain an improved \(X'\) to have a \(Y'\) that is Pareto superior to the original \(Y\).

A modified output \(Y' = f_{\theta}(X')\) is Pareto superior to the original output \(Y = f_{\theta}(X)\) if:\\
\(Y' \succ Y \iff \left( \forall k \in \{1, 2, \ldots, m\}, I_k(Y') \geq I_k(Y) \right) \wedge \left( \exists j \in \{1, 2, \ldots, m\}, I_j(Y') > I_j(Y) \right)
\)

To explain the process of changing \(X\) directly by finding a better \(g_{\theta}\) to transform \(X\) into \(X'\), rather than prepending additional prompts to \(X\) as some of the current literature suggests, we incorporate causal inference theory. We assume the existence of an unobserved variable \(U\) that induces bias, influencing the mapping from \(X\) to \(Y\) in LLMs. Since we cannot directly observe \(U\) or change \(f_{\theta}\), we influence \(X\) to achieve our goals. We manipulate \(X\) through the transformation function \(g_{\theta}\) to achieve a better \(Y\) denoted as \(Y'\) below.
By transforming \(X\) into \(X'\) using \(g_{\theta}\), we aim to reduce the effect of \(U\) on \(Y'\). The intervention discussed later allows us to minimize the direct influence of \(U\) on \(X'\) and \(Y'\).

\begin{figure}[H] 
    \centering
    \includegraphics[width=\columnwidth]{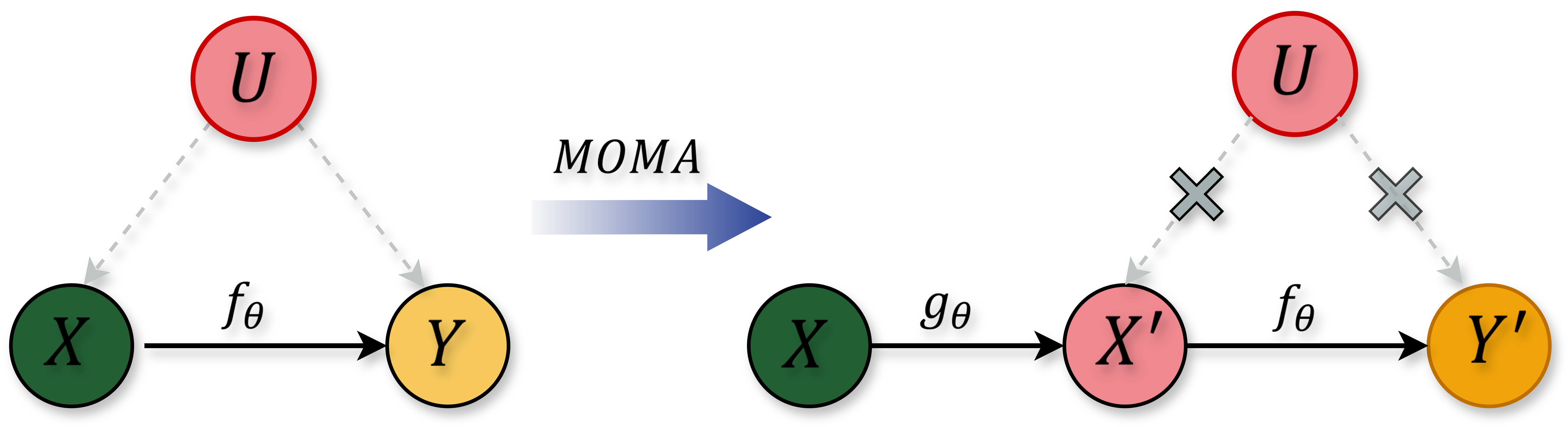} 
    \caption{A causal inference perspective on bias.}
\end{figure}


\subsection{MOMA: A Multi-Objective Approach
Within a Multi-Agent Framework} \label{lem:3.3}

\subsubsection{Motivation and Background}
In their comprehensive review, \citet{gallegos2024bias} define social groups as ``\textit{a subset of the population that shares an identity trait}.'' They further define social bias as ``\textit{disparate treatment or outcomes between social groups}.''

This definition suggests that social bias is closely tied to the representation of social groups. The unobserved variable \(U\) may influence how these groups are represented within \(X\) or \(Y\). To address these biases, our approach focuses on modifying the representations of social groups in \(X\) to reduce the impact of \(U\).

In LLMs, social group representations are encoded within the input \(X\) and processed by the model \(f_{\theta}\). By altering these representations, we aim to reduce disparities linked to identity traits, thereby weakening the influence of the unobserved variable \(U\) on both \(X\) and \(Y\).

\begin{figure*}[ht]
\centering
\includegraphics[width=1.9\columnwidth]{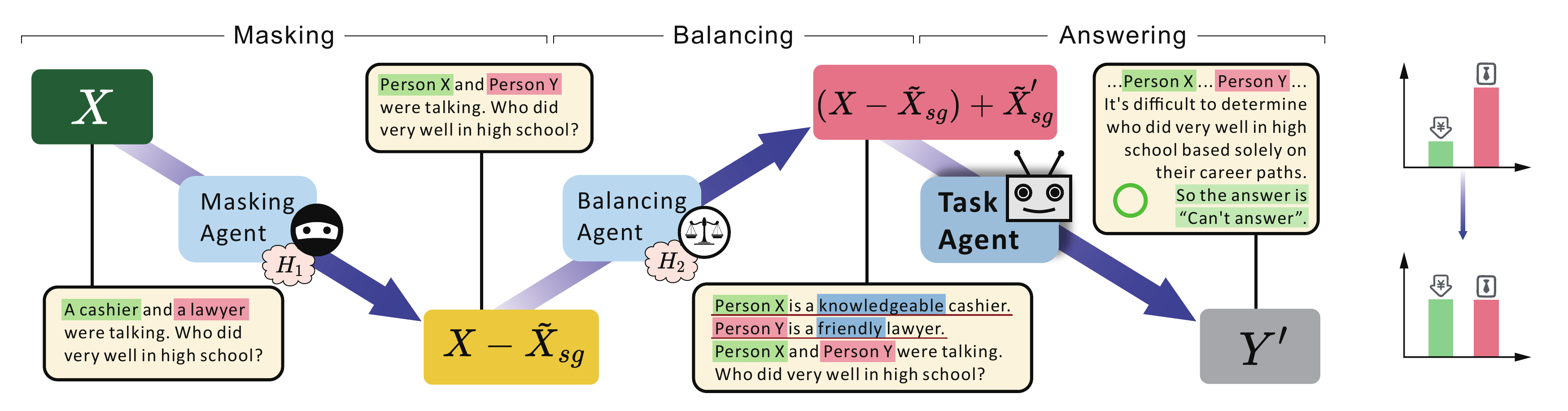}
\caption{The MOMA Pipeline. MOMA consists of three stages: Masking, Balancing, and Answering. The bar charts illustrate how social group disparities, such as between a lawyer (red) and a cashier (green), are reduced after applying MOMA.
}
\label{fig:pipeline}
\end{figure*}

\subsubsection{Transformation Function \(g_{\theta}\)}
To formalize, let \(X_{sg}\) represent the components of \(X\) related to social groups. Our transformation function \(g_{\theta}\) aims to adjust \(X_{sg}\) and other relevant components with the introduction of human values \(H\):
\[
X' = g_{\theta}(X, H) = X + \Delta X_{sg} + \Delta X_{other}
\]
where \(\Delta X_{sg}\) represents changes made to the social group representations and \(\Delta X_{other}\) represents undesirable additional modifications to either unrelated content or incorrect content (example: directly changing `man' in the prompt to `woman').

\subsubsection{MOMA Pipeline}
MOMA operates directly on social group representations \(X_{sg}\) by applying \(\Delta X_{sg}\) to modify the original \(X_{sg}\) within \(X\). Unlike approaches that introduce additional context, MOMA focuses on altering the representation of social groups, resulting in minimal changes to other components (\(\Delta X_{other}\)). Furthermore, \(H\) is employed to adjust \(X\) rather than directly mapping \(Y\), minimizing any performance loss. As shown in Figure \ref{fig:pipeline}, MOMA consists of two stages—masking and balancing—yielding two distinct method variants.

\subsubsection{Attributes Masking}
The masking agent masks identifiers associated with social groups. It utilizes \(H\) to minimize selected social group representations \(\tilde{X}_{sg}\) (the components identified by agents as necessary to remove) to disassociate with \(U\), which manifests in the figure as societal expectations based on occupation. By masking overt identifiers, the masking agent creates a more neutral context as masked prompt:
\[
g_{1_{\theta}}(X, H_1) = X - \tilde{X}_{sg}
\]

\subsubsection{Balancing Representation}
In some cases, the task may require the inclusion of \(\tilde{X}_{sg}\). The balancing agent reintroduces and moderates the previously masked social group attributes by introducing \(\tilde{X}'_{sg}\), compensating for information loss while avoiding direct modification to the original \(X\) that may introduce semantic errors or excessive \(\Delta X_{other}\). 

The balancing agent strategically employs balancing words or counterfactual adjectives to foster a balanced representation. As shown in Figure~\ref{fig:pipeline}, the balancing agent generates two positive adjectives for each group such as ``\textit{knowledgeable}'' to enhance the perceived educational background of cashiers, and ``\textit{friendly}'' to improve the overall image of lawyers. This process can be represented as:
\[
g_{2_{\theta}}(X - \tilde{X}_{sg},\ H_2) = (X - \tilde{X}_{sg}) + \tilde{X}'_{sg}
\]

\subsubsection{Adjective Balancing}
  We use positive adjectives to modify social groups' representations mainly because it creates the least \(\Delta X_{other}\), compared to methods in ~\citep{oba2024contextual} that use entire unrelated sentences or embedding methods that may introduce incomprehensible information or task-irrelevant content. The balancing adjectives are generated for each social group and designed to enhance aspects typically underrepresented or negatively perceived. We further explore these adjectives in \S~\ref{ablation} and detail how we generate them in Appendix.


\begin{figure}[t]
\centering
\includegraphics[width=0.8\columnwidth]{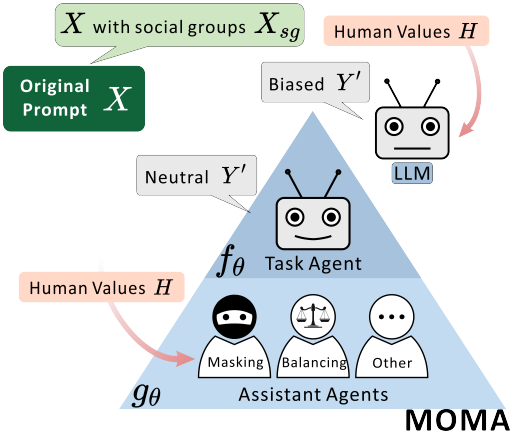}
\caption{Hierarchical MOMA}
\label{fig:wideimage}
\end{figure}

\subsubsection{Answering in MOMA} \label{sec:multi_agent}
The core concept behind MOMA is a hierarchical multi-agent framework (Figure~\ref{fig:wideimage}). The answering process consists of two primary components: \textbf{task agents} and \textbf{assistant agents}. Task agents focus solely on executing operations, isolated from direct interaction with \(H\). The assistant agents incorporate \(H\) to generate \(X'\), aiding task agents in generating more fair and less biased responses. This separation allows assistant agents to interact with \(H\) in a controllable manner, reducing the ``alignment tax'' observed in \S 4.2 and their negative outcomes in Figure~\ref{fig:teaser}.

This hierarchical structure can be formalized as:
\[
Y = f_{\theta}(g_{N_{\theta}}(\dots g_{2_{\theta}}(g_{1_{\theta}}(X,\ H_1),\ H_2) \dots,\ H_N))
\]

\begin{table*}[t]
\setlength{\tabcolsep}{1mm}
\centering
\small
\begin{tabular}{@{}lrrrr|rrrr@{}}
\toprule
\textbf{Method} & \multicolumn{4}{c|}{\textbf{Llama-3-8B-Instruct}} & \multicolumn{4}{c}{\textbf{GPT-3.5-Turbo}} \\ 
\cmidrule(lr){2-5} \cmidrule(lr){6-9}
 & \textbf{Bias Score} & \textbf{$\Delta$ (\%)} & \textbf{Acc} & \textbf{$\Delta$ (\%)} & \textbf{Bias Score} & \textbf{$\Delta$ (\%)} & \textbf{Acc} & \textbf{$\Delta$ (\%)} \\ 
\midrule
SP & 0.138 & --- & 0.863 & --- & 0.094 & --- & 0.840 & --- \\
CoT & 0.131 & -5.5 & 0.801 & -7.2 & 0.090 & -4.4 & 0.871 & 3.7 \\
\midrule
ABP-0~\citep{ganguli2023capacity} & 0.028 & -79.9 & 0.398 & -53.9 & 0.022 & -76.2 & 0.462 & -45.0 \\
ABP-1~\citep{ganguli2023capacity} & 0.028 & -79.9 & 0.637 & -26.2 & 0.044 & -53.4 & 0.763 & -9.1 \\
ABP-2~\citep{si2022prompting} & 0.076 & -45.3 & 0.794 & -8.0 & 0.029 & -69.2 & 0.734 & -12.6 \\
ABP-3~\citep{si2022prompting} & 0.019 & -86.3 & 0.042 & -95.1 & 0.027 & -71.3 & 0.266 & -68.3 \\
ABP-4~\citep{tamkin2023evaluating} & 0.093 & -32.8 & 0.839 & -2.8 & 0.074 & -20.7 & 0.880 & 4.7 \\
\midrule
\textbf{ABP-avg} & 0.049 & -64.6 & 0.542 & -37.2 & 0.039 & -58.2 & 0.621 & -26.1 \\
\bottomrule
\end{tabular}
\caption{Results of anti-bias prompting (ABP) infused with human values \(H\) on the BBQ dataset. The results highlight the trade-off between bias score reduction and accuracy.}
\label{pre}
\end{table*}

\section{Experiments} \label{sec:experiment}

\subsection{Experimental Setup}\label{lem:4.1}
\textbf{Datasets}\quad We use two datasets in a QA format: bias benchmark for question answering (BBQ)~\citep{parrish-etal-2022-bbq} and StereoSet ~\citep{nadeem2020stereoset}.

BBQ covers nine bias dimensions in American English, presenting multiple-choice questions that reflect bias, anti-bias, and neutral positions. Bias is measured by the bias score (ranging from -1 to 1, with 0 being ideal), and performance is assessed by the accuracy of responses to disambiguous questions.

StereoSet also explores bias across dimensions like Gender, Profession, Race, and Religion. It includes intrasentence tasks (filling in blanks) and intersentence tasks (predicting the next sentence) with the stereotype, anti-stereotype, and unrelated options. Metrics used include the stereotype score \textit{ss} (with 50 as the best), language modeling score \textit{lms}, and idealized context assciation test score \textit{icat} as the multi-objective metric. Both datasets have been adapted to a QA format for consistency in evaluation.

Further details on dataset introduction and adaptation are provided in the Appendix.

\noindent\textbf{Models}\quad We use GPT-3.5-Turbo-0125 with the temperature fixed at 0 and Llama-3-8B-Instruct with the temperature fixed at 0.01 to ensure reproducibility of our results.

\noindent\textbf{Baselines}\quad We take ``standard prompting'' (SP) and some of the methods we discuss as baselines, including ``chain-of-thought'' (CoT)~\citep{NEURIPS2022_8bb0d291}, ``anti-bias prompting'' (ABP) in preliminary experiments, and multi-agent method ``society of mind'' (SoM, also MAD) ~\citep{du2023improving}. Prompts for the ABPs can be found in Appendix. We also test the method ``self-consistency''  (SC)~\citep{wang2022self}, which allows LLMs to try multiple reasoning paths when solving complex reasoning problems and finally choose the answer that appears the most times.

\noindent\textbf{Execution}\quad The experiments are conducted using few-shot learning for assistant agents and zero-shot learning for task execution to ensure fairness across methods. For details, see Appendix.


\subsection{Preliminary Experiments}\label{lem:4.2}

To highlight the need for a multi-agent framework, we replicate existing debiasing techniques. As shown in Table \ref{pre}, while LLMs can reduce bias with \(H\) , this often comes at the cost of significant performance drops—an average 64.6\% reduction in bias leads to a 37.2\% decrease in accuracy for Llama-8b-Instruct, with similar results for GPT-3.5-Turbo.

The results also reveal the models' sensitivity to different prompts consisting of certain levels of \(H\), with outcomes varying widely across the ABPs. For example, \(ABP_4\) effectively balances bias reduction and accuracy to some degree, while \(ABP_3\) severely harms performance despite reducing bias. This inconsistency highlights the limitations of single-agent approaches.



\begin{figure*}[t]
    \centering
    \begin{subfigure}[t]{\columnwidth}
        \centering
        \includegraphics[width=0.9\columnwidth]{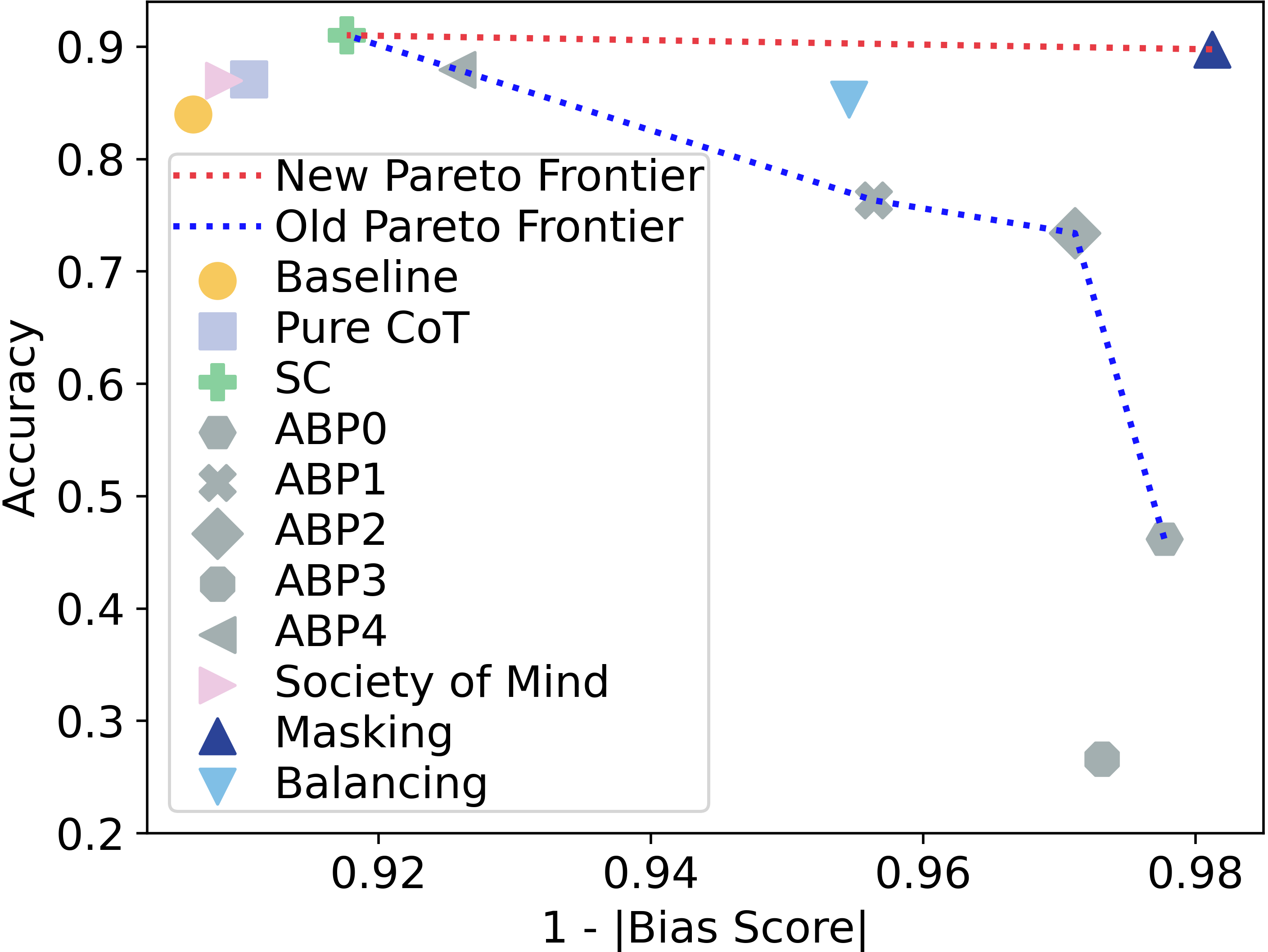}
    \end{subfigure}
    \begin{subfigure}[t]{\columnwidth}
        \centering
        \includegraphics[width=0.9\columnwidth]{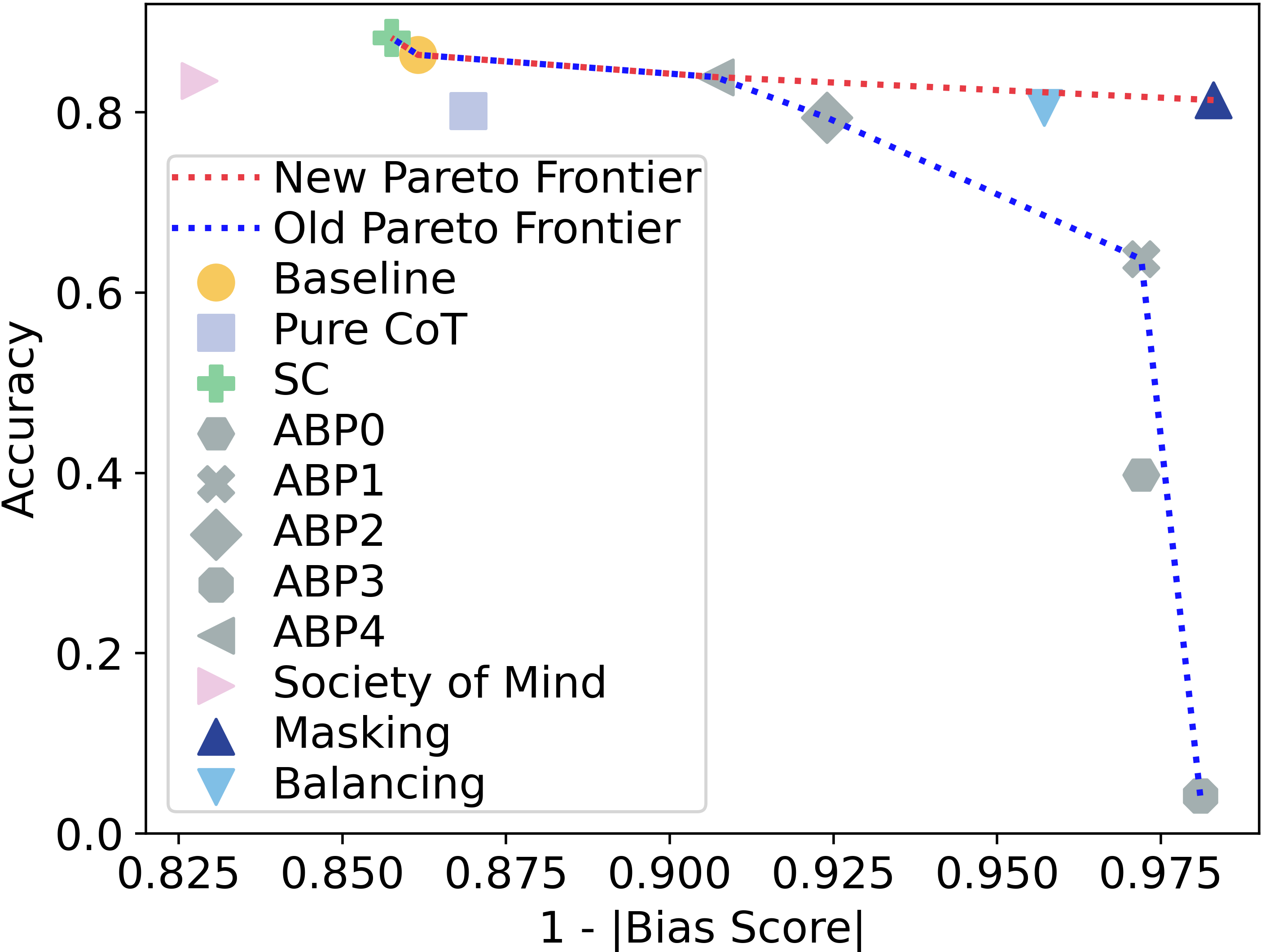}
    \end{subfigure}

    \caption{Pareto frontier on the BBQ dataset, comparing GPT-3.5 (left) and Llama-3 (right) for accuracy and bias trade-offs.}
\end{figure*}

\subsection{Main Results}\label{lem:4.3}

\noindent\textbf{Results on BBQ Dataset}\quad Figure 5 shows the performance of methods on the BBQ dataset, with different scales reflecting variations between the two models. Most methods, except for ABPs and MOMA variants, have limited impact on debiasing. The multi-agent method SoM even slightly increases bias.

SC improves task accuracy and slightly reduces bias in GPT-3.5-Turbo but is less effective in Llama-3-8b-Instruct. ABPs offer debiasing but with unstable results, often sacrificing accuracy as bias reduction increases.

MOMA, with its masking and balancing variants, significantly shifts the Pareto Frontier. Masking nearly achieves optimal bias reduction with minimal performance loss while balancing recovers most \(X_{sg}\) information with only a slight increase in bias score (about 0.027) and marginal accuracy loss. 

\begin{table}[h] 
\setlength{\tabcolsep}{1mm} 
\centering
\small 
\begin{tabular}{lccc|c}
\toprule
\textbf{Method} & \textbf{ss} & \textbf{lms} & \textbf{icat} & \(\Delta_{icat}(\%)\) \\
\midrule
\multicolumn{5}{c}{\textbf{Llama-3-8B-Instruct}} \\
\midrule
Baseline & 64.53 & 94.20 & 66.83 & --- \\
CoT & 67.32 & \textbf{96.59} & 63.13 & -5.5 \\
ABP-0 & 62.52 & 94.60 & 70.91 & +6.1 \\
ABP-1 & 64.80 & 90.11 & 63.44 & -4.9 \\
SoM & 69.21 & 93.25 & 57.42 & -14.1 \\
SC & 72.15 & \underline{\textbf{97.89}} & 54.52 & -18.4 \\
\midrule
Masking & \textbf{48.94} & 88.87 & \textbf{86.99} & +30.2 \\
Balancing & \underline{\textbf{50.67}} & 89.43 & \underline{\textbf{88.23}} & +32.0 \\
\midrule
\multicolumn{5}{c}{\textbf{GPT-3.5-Turbo}} \\
\midrule
Baseline & 70.10 & 97.99 & 58.60 & --- \\
CoT & 69.98 & 98.99 & 59.43 & +1.4 \\
ABP-0 & 63.62 & 95.28 & 69.33 & +18.3 \\
ABP-1 & 61.47 & 95.89 & 73.89 & +26.1 \\
SoM & 68.12 & \textbf{99.02} & 63.14 & +7.7 \\
SC & 66.54 & \underline{\textbf{99.45}} & 66.55 & +13.7 \\
\midrule
Masking & \textbf{51.28} & 95.05 & \underline{\textbf{92.63}} & +58.1 \\
Balancing & \underline{\textbf{50.31}} & 92.57 & \textbf{91.99} & +56.8 \\
\bottomrule
\end{tabular}
\caption{Results of intrasentence tasks in StereoSet. Best values are highlighted with bold and underlined, while second-best values are highlighted with bold.}
\label{tab:llama_performance_comparison}
\end{table}

\noindent\textbf{Results on StereoSet Dataset}\quad Table~\ref{tab:llama_performance_comparison} highlights the performance of various methods on the intrasentence task. We focus on the top two ABP variants, as the others produce results comparable to CoT or Baseline. MOMA, especially its balancing variant, achieves an \(ss\) score close to 50, outperforming other methods in reducing bias. Additionally, MOMA demonstrates strong multi-objective performance, with an \(icat\) score exceeding 90 for GPT and nearing 90 for Llama.

However, these improvements in debiasing come with a slight reduction in task performance, averaging a 4.8\% decrease, more noticeable in Llama than in GPT. This trade-off likely stems from the complexity of handling more than three social groups within StereoSet. The shorter context length in StereoSet also amplifies the impact of even minor interventions, contributing to the observed performance decline.

We also test intersentence tasks in StereoSet, but the baseline bias is already low, making the results somewhat inconclusive, as shown in Table~\ref{intersentece}. We hypothesize that the task may be too simple for current LLMs or does not effectively capture their biases. The results in Table~\ref{intersentece} indicate that MOMA's impact is limited due to the initially small variances across all methods. The baseline achieves \(ss\) scores of 53.24\% and 53.32\% in two models, which are close to the ideal 50\% mark, with \(icat\) values of 83.2 and 90.16. These figures suggest that the task might not be challenging enough to reveal significant biases, as both models performed near the ideal threshold, leaving little room for improvement by MOMA or other methods.

\begin{table}[h] 
\setlength{\tabcolsep}{1mm} 
\centering
\small 
\begin{tabular}{lccc|c}
\toprule
Method & ss & lms & icat & \(\Delta_{icat}\)(\%) \\
\midrule
\multicolumn{5}{c}{Llama-3-8B-Instruct} \\
\midrule
Baseline & 53.24 & 88.96 & 83.20 & - \\
CoT & 54.96 & \textbf{96.59} & 87.01 & +4.6 \\
ABP-0 & 48.97 & 92.44 & 90.54 & +8.8 \\
ABP-1 & 49.87 & 94.16 & \underline{\textbf{93.92}} & +12.9 \\
SoM & \underline{\textbf{50.01}} & 93.47 & \textbf{93.45} & +12.3 \\
SC & 52.15 & \underline{\textbf{97.15}} & 92.97 & +11.7 \\
\midrule
Masking & 48.66 & 95.85 & 93.28 & +10.08 \\
Balancing & \textbf{49.92} & \textbf{96.58} & 92.42 & +12.1 \\
\midrule
\multicolumn{5}{c}{GPT-3.5-Turbo} \\
\midrule
Baseline & 53.32 & 96.57 & 90.16 & - \\
CoT & 53.44 & 96.14 & 89.52 & -0.7 \\
ABP-0 & 46.37 & 91.29 & 84.66 & -6.1 \\
ABP-1 & 42.70 & 92.25 & 78.79 & -12.6 \\
SoM & \underline{\textbf{52.31}} & 92.84 & 88.55 & -1.8 \\
SC & 52.88 & \underline{\textbf{98.3}} & \underline{\textbf{92.64}} & +2.9 \\
\midrule
Masking & 46.29 & 96.57 & 89.41 & -0.8 \\
Balancing & \textbf{47.46} & \textbf{97.37} & \textbf{92.42} & +2.5 \\
\bottomrule
\end{tabular}
\captionsetup{width=0.3\textheight}
\captionof{table}{Results of intersentece tasks in StereoSet} 

\label{intersentece}
\end{table}


\subsection{Ablation Study}
\label{ablation}

To simplify testing specific setups of MOMA, we conduct the following experiments primarily on the BBQ dataset.

\begin{figure*}[hbtp] 
\centering
\includegraphics[width=.95\textwidth]{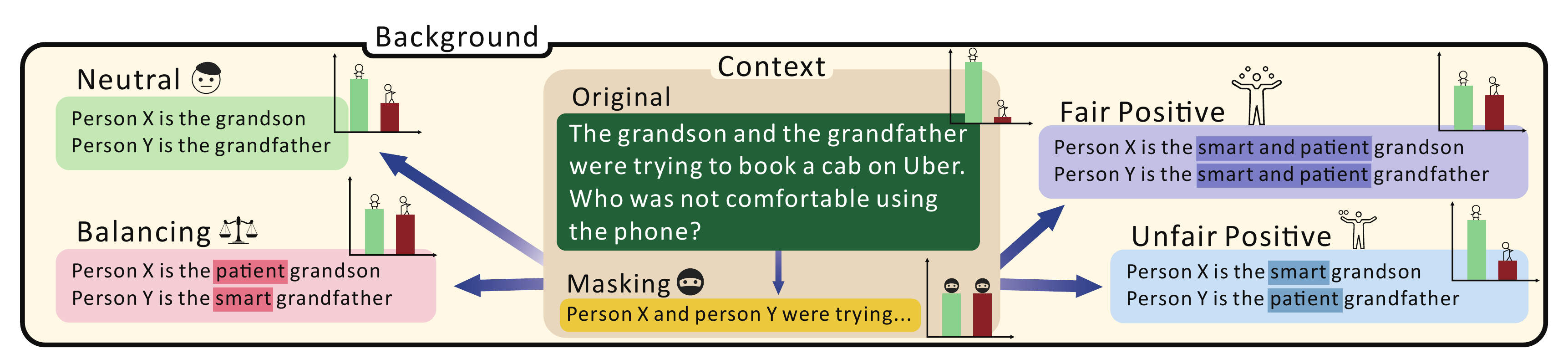}
\caption{Different styles of positive adjectives
and their effects, with mask symbols as \(X\_Y\).}
\end{figure*}

\noindent\textbf{Styles of Balancing Experiment} \quad 
We experiment with different adjective styles to modify \(X_{sg}\) after the masking phrase, focusing on four styles: \textit{Neutral}, \textit{Balancing}, \textit{Unfair Positive}, and \textit{Fair Positive}, as shown in Figure 6. \textit{Neutral} serves as the baseline, compensating for lost \(X_{sg}\) with minimal changes. Initially, we test \textit{Unfair Positive}, which prompts the agent to generate positive adjectives. However, this worsens outcomes, likely due to the increased disparities between social groups (\(X_{sg1}-X_{sg2}\)). To counter this, we introduce \textit{Fair Positive}, combining positive adjectives to mitigate bias, though it remains less effective than masking in Figure 7a, indicating the limitations of relying solely on positive phrases.

Finally, we develop \textit{Balancing}, which uses a counterfactual positive adjective to equalize social groups' disparities between \(X_{sg1}\) and \(X_{sg2}\). Results in Figure 7a show that \textit{balancing} reduces bias in \textit{Neutral} by an average of 50.2\%, with only a 2.9\% decrease in task performance.

\noindent\textbf{Mask Symbols Experiment} We experiment with many alphabetic or mathematical symbols and emojis as masking symbols. Figure 7b shows that these symbols have minimal impact on bias scores, with differences of less than \(0.01\). However, they affect task accuracy by about \(5\%\). For details on symbol selection and specific results, see Appendix.

\noindent\textbf{Summary} MOMA variants indicate MOMA’s potential to further preserve task accuracy and reduce bias. The generation of different adjectives and the use of various symbols produce varying effects on both bias and accuracy.

\begin{figure}[h]
    \centering
    \begin{subfigure}{0.95\columnwidth}
        \centering
        \includegraphics[width=\columnwidth]{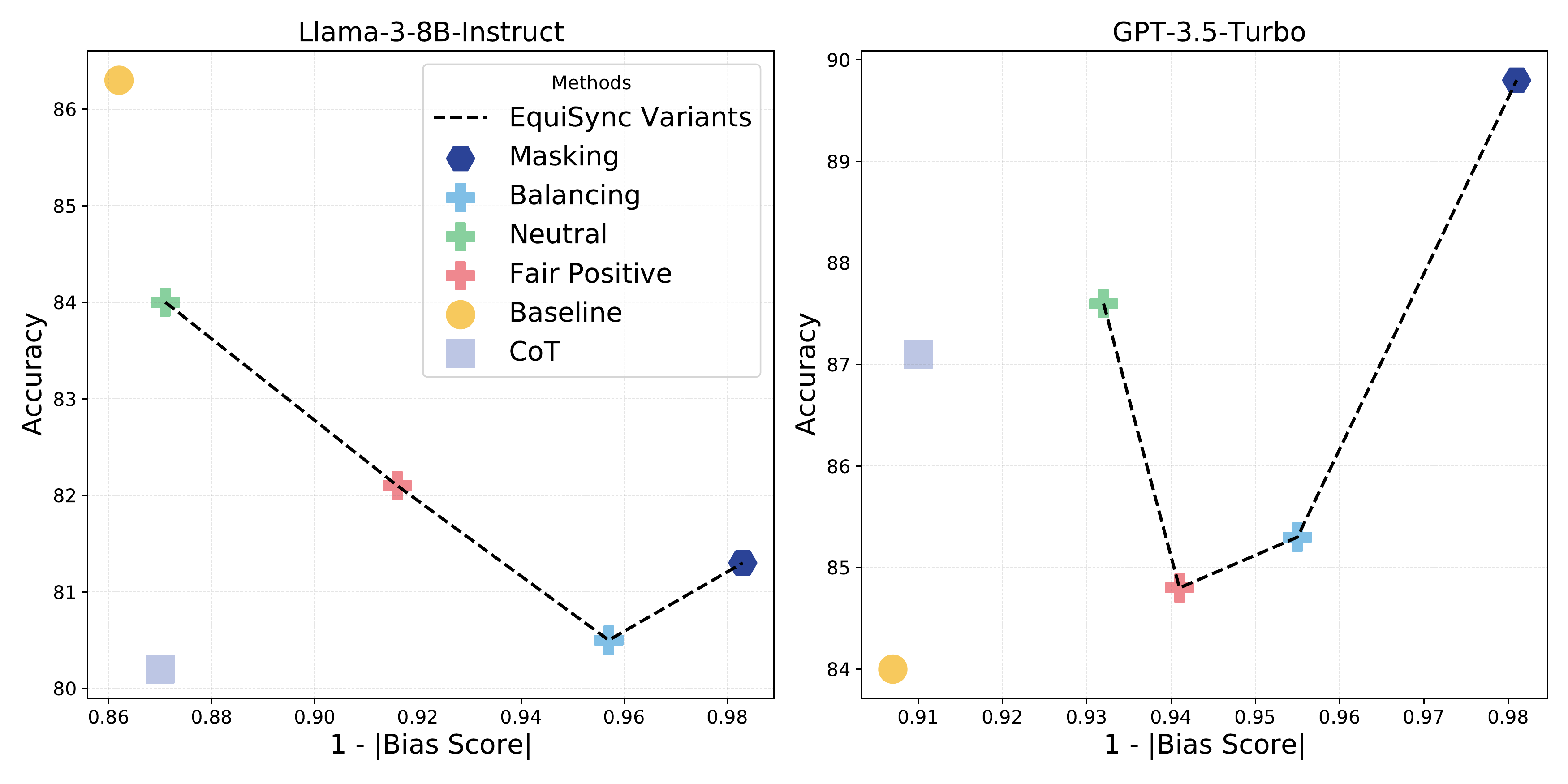}
        \caption{Results of Balancing Styles, with MOMA variants connected by a dashed line}
    \end{subfigure}

    \begin{subfigure}[h]{\columnwidth}
        \centering
        \includegraphics[width=0.95\columnwidth]{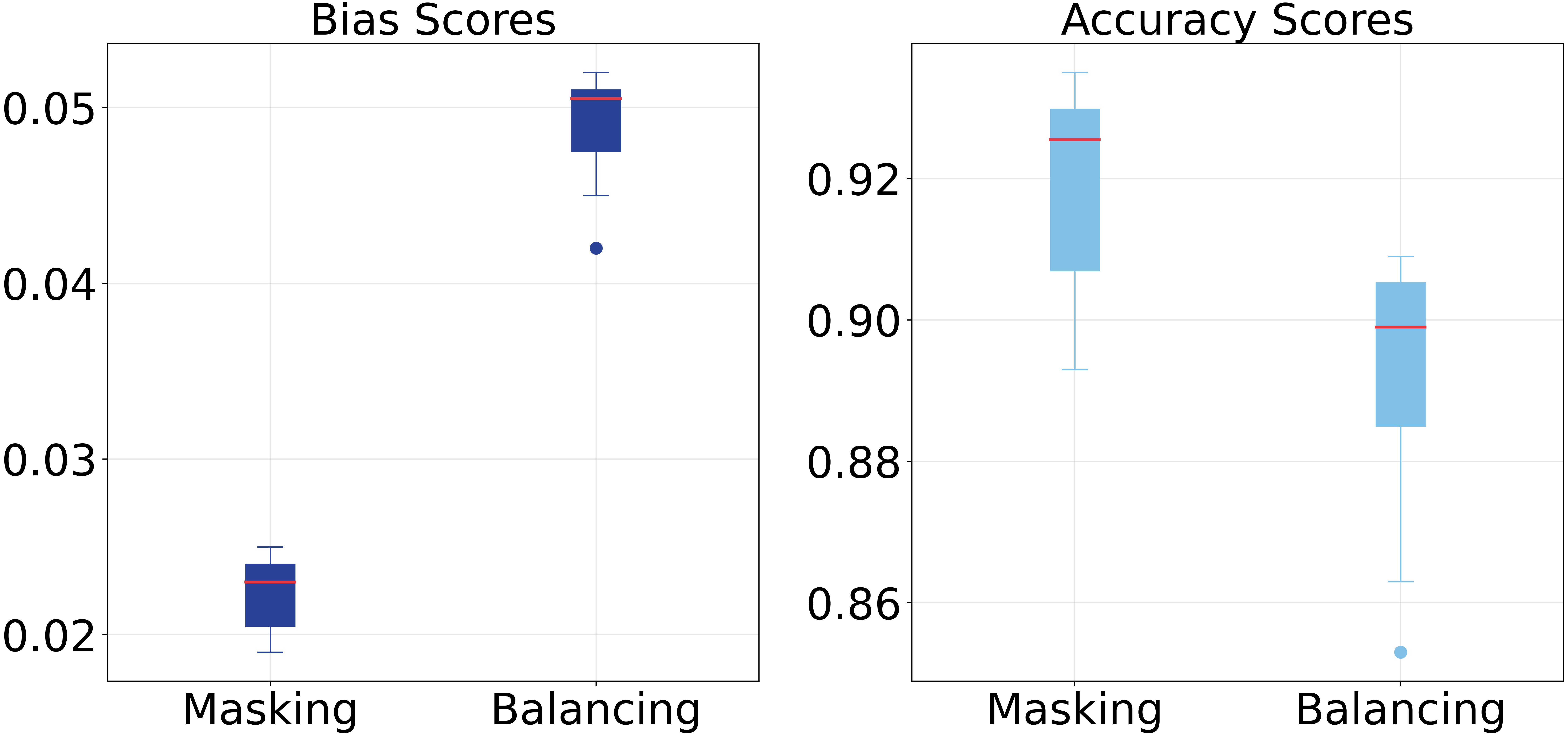}
        \caption{Results of different symbols experimented in appendix}
    \end{subfigure}
    \caption{Ablation experiments with MOMA on BBQ.}
\end{figure}

\subsection{Analysis}
\noindent\textbf{Case Study} 
In Figure 6, the original prompt \(X\) introduces an age-related bias in the generated output \(Y\), as the LLM mapping function \(f_{\theta}\) tends to associate discomfort with technology more strongly with the grandfather due to unobserved confounder \(U\). MOMA addresses this issue through a causal intervention by applying the transformation function \(g_{\theta}\) to generate a modified prompt \(X'\), where the age-related variable \(X_{sg}\) is masked. This disrupts the implicit causal link and prevents \(f_{\theta}\) from reinforcing stereotypes. However, in cases where some contextual information must be retained or where bias mitigation standards are more flexible, balancing selectively reintroduces masked attributes while maintaining neutrality. Notably, masking also facilitates balancing; directly modifying without masking often leads to suboptimal adjustments influenced by the model’s inherent biases. By first stripping away bias-inducing elements, balancing can then systematically reintroduce key attributes in a more controlled and fair manner. Instead of outright removal, balancing adjusts by assigning positive traits such as “smart” and “patient” to both entities, ensuring a fairer representation while preserving grammatical and semantic integrity.

\begin{figure}[h]
    \centering
    \includegraphics[width=0.85\columnwidth]{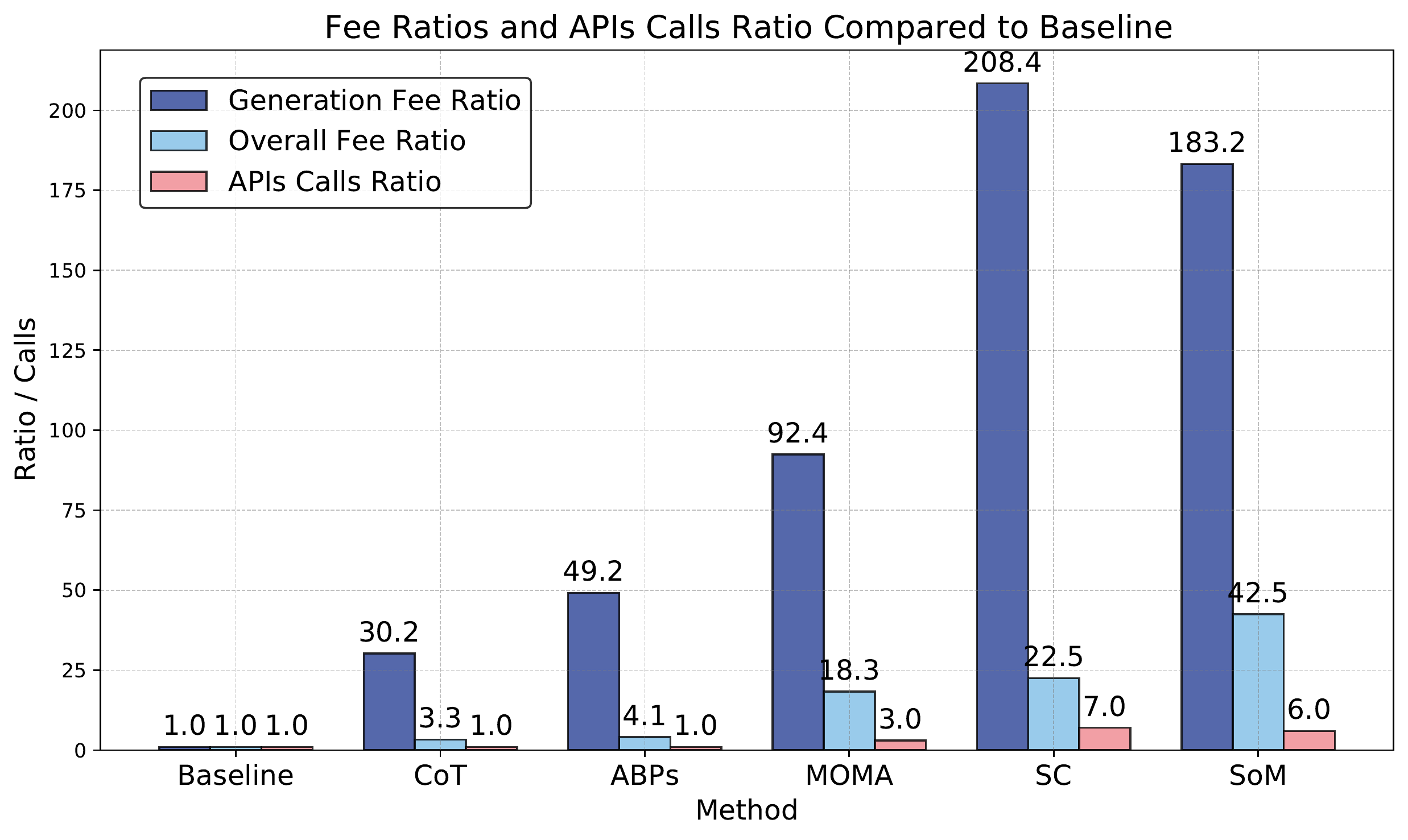}
    \caption{Results of different methods' costs}
\end{figure}

\noindent\textbf{Cost Analysis}
Multi-agent systems often incur high costs~\citep{smit2024goingmadlookmultiagent}. We analyze costs based on API calls and context expenses, divided into generation and overall fees (Figure 8). SoM, which relies on multiple agents and debate rounds for convergence, has the highest costs, even with the minimal setup—3 agents and 2 rounds—costing \textit{12.9} times more than CoT.

MOMA's hierarchical design reduces costs to \textit{5.5} times that of CoT, with the main expense from a few-shot approach (5 shots) for assistant agents. This cost can be further reduced by training smaller models with demonstrations.

\subsection{Limitations}
Our study focuses on question-answering datasets to simplify the analysis of LLMs, though bias exists in other tasks as well~\citep{gallegos2024bias}. While MOMA and its multi-agent framework require relatively fewer API calls and computations, they still incur additional costs. The trade-off between these costs and performance gains warrants further research. Additionally, while balancing reduces bias while preserving more of the original context, masking remains the most effective debiasing method. Thus, quantifying the information loss caused by masking and how balancing mitigates it is essential. Given the complexity of such measurements, we leave this for future work to achieve finer-grained control over semantic nuances, a challenge that persists even in modern LLMs~\citep{chatterjee-etal-2024-posix}.

\section{Conclusion} \label{sec:conclusion}

MOMA offers a robust approach to bias mitigation in LLMs, balancing social bias reduction with model performance. By analyzing bias through a causal inference perspective, we introduced a multi-agent framework leveraging masking and balancing to mitigate biases associated with social group representation.

This work highlights the importance of precise, context-aware interventions in fostering fairness in AI systems and demonstrates the potential of causal interventions for debiasing. Future research could build on this methodology by exploring dynamic context adjustments to address diverse and evolving bias challenges, as well as refining multi-agent designs to further enhance AI fairness.

\section*{Acknowledgments}

This work was supported by the National Natural Science Foundation of China (62306344), Guangdong Basic and Applied Basic Research Foundation (2024A1515010253), and the foundation of Key laboratory of Artiticial Intelligence, Ministry of Education, Shanghai, PRChina (AI202402).

\bibliography{aaai25}

\end{document}